\documentclass[runningheads]{llncs}
\usepackage{tabularx}
\usepackage{subfigure}
\usepackage{amssymb}

\usepackage{epsfig}
\usepackage{url}
\usepackage{extarrows}
\usepackage{booktabs}
\begin{document}
\title{Multi-Representation Adaptation Network for Cross-domain Image Classification}
\author{Yongchun Zhu\inst{1,2} \and
Fuzhen Zhuang\inst{1,2}\thanks{Corresponding author: Fuzhen Zhuang} \and
Jindong Wang\inst{3} \and
Jingwu Chen\inst{5} \and
Zhiping Shi\inst{4} \and
Wenjuan Wu\inst{6} \and
Qing He\inst{1,2}
}
\authorrunning{Zhu et al.}
\institute{$^1$Key Lab of Intelligent Information Processing of Chinese Academy of Sciences (CAS), Institute of Computing Technology, CAS, Beijing 100190, China \\
$^2$University of Chinese Academy of Sciences, Beijing 100049, China \\
$^3$Microsoft Research,  $^4$Capital Normal University, $^5$ByteDance \\
$^6$School of Information, Renmin University of China, Beijing 100872, China 
\email{\{zhuyongchun18s,zhuangfuzhen\}@ict.ac.cn}}
\maketitle              % typeset the header of the contribution
\begin{abstract}
In image classification, it is often expensive and time-consuming to acquire sufficient labels. To solve this problem, domain adaptation often provides an attractive option given a large amount of labeled data from a similar nature but different domain. Existing approaches mainly align the distributions of representations extracted by a single structure and the representations may only contain partial information, e.g., only contain part of the saturation, brightness, and hue information. Along this line, we propose Multi-Representation Adaptation which can dramatically improve the classification accuracy for cross-domain image classification and specially aims to align the distributions of multiple representations extracted by a hybrid structure named Inception Adaptation Module (IAM). Based on this, we present Multi-Representation Adaptation Network (MRAN) to accomplish the cross-domain image classification task via multi-representation alignment which can capture the information from different aspects. In addition, we extend Maximum Mean Discrepancy (MMD) to compute the adaptation loss. Our approach can be easily implemented by extending most feed-forward models with IAM, and the network can be trained efficiently via back-propagation. Experiments conducted on three benchmark image datasets demonstrate the effectiveness of MRAN. The code has been available at \url{https://github.com/easezyc/deep-transfer-learning}.

\end{abstract}

\section{Introduction}\label{sec:intro}
As one of the fundamental technologies in computer vision, image classification has widely been researched. There are many applications of image classification, such as face recognition~\cite{parkhi2015deep}, handwritten recognition~\cite{lecun1990handwritten}, and human activity recognition~\cite{bulling2014tutorial}. To successfully construct an image classification system, a sufficient number of manually annotated images for each specific target domain are required beforehand. With a large amount of labeled training data and substantial computation resources, the satisfying performances can be achieved by deep neural networks recently~\cite{simonyan2015very,he2016deep}. 

Nevertheless, in real situations, it is usually impractical to obtain sufficient manually labeled training data for every new scenario. Moreover, it is often prohibitively difficult and expensive to obtain enough labeled data. To alleviate this problem, domain adaptation~\cite{pan2010survey,zhuang2015supervised}, which aims to adapt the feature representation learned in the source domain with rich label information to the target domain with less or even no label information, has received much attention in recent years.

Recent domain adaptation methods achieve remarkable results by embedding domain adaptation modules in the pipeline of deep feature learning to extract domain-invariant representations. This can generally be achieved by optimizing some measures of domain shift~\cite{quionero2009dataset,pan2010survey}, e.g., maximum mean discrepancy~\cite{tzeng2014deep,long2015learning}, correlation distances~\cite{sun2016return,sun2016deep}, or minimizing an approximate domain discrepancy distance through an adversarial objective with respect to a domain discriminator~\cite{ganin2015unsupervised,tzeng2017adversarial}.

\begin{figure}[t!]
\centering
\begin{minipage}[b]{1\linewidth}
\centering
\includegraphics[width=1\linewidth]{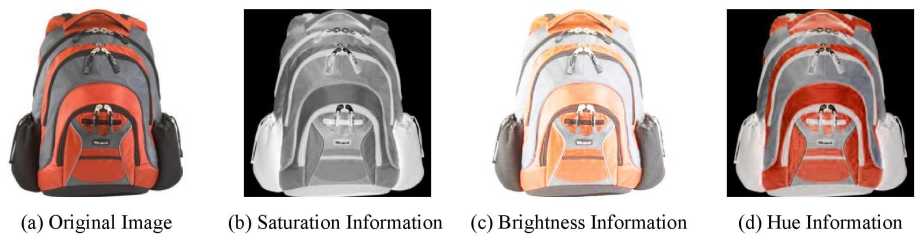}
\end{minipage}
\caption{(a) is the original image, while the parts of information in (b), (c), (d) are captured from (a) by different structures. (b), (c), (d) only contain part of the saturation, brightness and hue information, respectively. (The original image is from the Office31 dataset~\cite{saenko2010adapting}.)}\label{fig1}
\end{figure}

Most of the recent deep domain adaptation methods are based on convolutional neural networks which have the ability to extract abstract representations from high-dimensional images. However, this feature extraction process might lose some important information. Hence, comparing to the original images, the representations may only contain partial instead of complete information, e.g., only contain part of the saturation, brightness, and hue information. An intuitive example is given in Figure~\ref{fig1}. Figure~\ref{fig1}(a) is the original image, while Figures~\ref{fig1}(b)$\sim$(d) are the transformed forms. More importantly, we find that all transformed images only contain partial information, thus they may give us wrong or distorted facts of the real image. Therefore, we need to observe the objects from multiple points to get a comprehensive understanding.

% \begin{figure*}[t!]
% \centering
% \subfigure[Original Image]{
% \begin{minipage}[b]{0.2\linewidth}
% \centering
% \includegraphics[width=1.\columnwidth,height=1.\columnwidth]{figure/original.eps}
% \label{fig1:a}
% \end{minipage}
% }
% \subfigure[Saturation Information]{
% \begin{minipage}[b]{0.2\linewidth}
% \centering
% \includegraphics[width=1.\columnwidth,height=1.\columnwidth]{figure/saturation.eps}
% \label{fig1:b}
% \end{minipage}
% }
% \subfigure[Brightness Information]{
% \begin{minipage}[b]{0.2\linewidth}
% \centering
% \includegraphics[width=1.\columnwidth,height=1.\columnwidth]{figure/brightness.eps}
% \label{fig1:c}
% \end{minipage}
% }
% \subfigure[Hue Information]{
% \begin{minipage}[b]{0.2\linewidth}
% \centering
% \includegraphics[width=1.\columnwidth,height=1.\columnwidth]{figure/hue.eps}
% \label{fig1:d}
% \end{minipage}
% }
% \caption{(a) is the original image, while the parts of information in (b), (c), (d) are captured from (a) by different structures. (b), (c), (d) only contain part of the saturation, brightness and hue information, respectively. (The original image is from the Office31 dataset~\cite{saenko2010adapting}.)}\label{fig1}
% \end{figure*}

The previous deep domain adaptation methods mainly align the distributions from representations extracted from the source and target domain data by a single structure, i.e., they are single-representation adaptation methods. Similar to the transformed images in the example, the representations only contain partial information, hence the alignment also focuses on partial information. Hence, this might lead to unsatisfying transfer learning performance. To fully understand the objects, more representations should be considered when aligning the distributions. To this end, different structures of convolutional neural networks provide an option to extract multiple representations from images. Along this line, we propose Multi-Representation Adaptation (MRA), which tries to align the distributions of the source and target domains using multiple representations extracted by a hybrid neural structure.

Specifically, we propose a Multi-Representation Adaptation Networks (MRAN) to align distributions of multiple representations in a domain-specific layer across domains for unsupervised domain adaptation. To enable MRA, we propose a hybrid neural structure named Inception Adaptation Module (IAM) to extract multiple representations from images. A key novelty over previous single-representation adaptation methods is the capability of MRAN to learn multiple domain-invariant representations which contain more information. Furthermore, the nonparametric Maximum Mean Discrepancy~\cite{gretton2012kernel} (MMD) is extended to compute the adaptation loss based on conditional distribution, and integrated into deep neural networks. The IAM method can be implemented by most feed-forward models and trained efficiently using standard back-propagation. Extensive experiments performed on three benchmark datasets show that MRAN can achieve remarkable performance compared with state-of-the-art competitors.

The contributions of this paper are summarized as follows. (1) To the best of our knowledge, we are the first to learn multiple different domain-invariant representations by Inception Adaptation Module (IAM) for cross-domain image classification. (2) A novel Multi-Representation Adaptation Network (MRAN) is proposed to align distributions of multiple different representations which might contain more information about the images. (3) MMD is extended to measure the discrepancy of conditional distributions across different domains in deep neural networks. (4) Finally, we conduct extensive experiments to validate the effectiveness of MRAN.

\section{Related Work}\label{sec:relatedWork}
Our work mainly belongs to domain adaptation, and we will introduce the related work in three aspects: image classification, domain adaptation, and multi-view learning.

\textbf{Image Classification}. As one of the fundamental technologies in computer vision, image classification has widely been researched. On the basis of the assumption of the parameter on data, the image classifiers could be divided into parametric and non-parametric classifier. For parametric classifier, the parameters like mean vector and covariance matrix are frequently generated from training samples, such as Maximum likelihood, linear discriminant analysis. While non-parametric classifiers do not make use of statistical parameters to calculate class separation, such as neural network, svm, decision tree. Recently, the deep neural networks~\cite{simonyan2015very,he2016deep,wang2018scene,wang2018locality} have achieved remarkable performance for image classification. With the guidance of the human visual system (HVS), Wang et al.~\cite{wang2018scene} explore the attention mechanism and propose a novel endto-end attention recurrent convolutional network (ARCNet) for scene classification. LSLRR~\cite{wang2018locality} improves the classical low-rank representation with locality constraint criterion and structure preserving
strategy. However, they assume the training and test sets have the same distributions. Hence, these methods could not solve the cross-domain problems.

\textbf{Domain Adaptation}. Recent years have witnessed many approaches to solve the visual domain adaptation problem, which is also commonly framed as the visual dataset bias problem~\cite{quionero2009dataset,pan2010survey,zhuang2015survey}. Previous shallow methods for unsupervised adaptation include re-weighting the training data so that they could more closely reflect those in the test distribution~\cite{jiang2007instance}, and finding a transformation in a lower-dimensional manifold that draws the source and target subspaces closer~\cite{gong2012geodesic,pan2011domain}. 

Most existing methods learn a shallow representation model to minimize domain discrepancy, which can not suppress domain-specific explanatory factors of variations. Deep networks learn abstract representations that disentangle the explanatory factors of variations behind data~\cite{bengio2013representation} and extract transferable factors underlying different populations~\cite{glorot2011domain,oquab2014learning}. Thus deep neural networks have been explored for domain adaptation~\cite{tzeng2014deep,ganin2015unsupervised,long2015learning,zhuang2015supervised,sun2016deep,tzeng2017adversarial,zhu2019adaptively}, where significant performance gains have been witnessed compared to prior shallow transfer learning methods. 

The main strategy of deep transfer networks is to guide feature learning by minimizing the difference between the source and target distributions.
Some recent works bridge deep learning and domain adaptation~\cite{long2015learning,long2016deep,ganin2015unsupervised,pei2018multi,tzeng2015simultaneous,tzeng2017adversarial}, which extend deep convolutional neural networks (CNNs) to domain adaptation. These works are mainly divided into two class, embedding methods by adding adaptation layers through which the embedding of distributions are matched~\cite{tzeng2014deep,long2015learning,long2016deep,zhu2019aligning}, and adversarial methods by adding a subnetwork as domain discriminator while the deep features are learned to confuse the discriminator in a domain-adversarial training paradigm~\cite{bousmalis2017unsupervised,ganin2016domain,hoffman2017cycada,liu2016coupled,saito2018maximum,tzeng2017adversarial,zhang2018collaborative,kang2018deep,kumar2018co}. And recent related work extends the adversarial methods to a generative adversarial way~\cite{bousmalis2017unsupervised}. Besides of these two mainstreams, there are diverse methods to learn domain-invariant features: DRCN~\cite{ghifary2016deep} reconstructs features to images and makes the transformed images are similar to original images. D-CORAL~\cite{sun2016deep} ``recolors" whitened source features with the covariance of features from the target domain. All of these methods focus on aligning distributions of representations extracted by a single structure. However, the representations might only contain partial information. Our MRA could cover more information by aligning distributions of multiple representations extracted by a hybrid structure. Therefore, the representation capability can be enhanced.

\textbf{Multi-view learning} is concerned with the problem of machine learning from data represented by multiple distinct feature sets. The recent emergence of this learning mechanism is largely motivated by the property of data from real applications where examples are described by different feature sets or different `views'. And multi-view learning arouses
amounts of interests in the past decades~\cite{blum1998combining,gonen2011multiple,yarowsky1995unsupervised}. Different from multi-view learning which needs data represented by multiple distinct feature sets, Multi-Representation learning focuses on extracting the multiple representations from the single view of data by a hybrid structure.
%%%
%%%\textit{Ensemble learning} typically refers to methods that generate several models which are combined to make a prediction, either in classification or regression problems. There has been large amount of research in recent years and good results have been reported (e.g.,~\cite{liu2000evolutionary, breiman2001random, rodriguez2006rotation}). The advantage of ensembles with respect to single models has been reported in terms of increased robustness and accuracy~\cite{garcia2005cooperative}. Motivated by that combination could increase robustness and accuracy, Multi-Representation Adaptation combines multiple representations. However, different from ensemble learning which cost much time to train several models, Multi-Representation Adaptation just need to train single model. 
\section{Multi-Representation Adaptation Networks} 
\label{sec:model}
In unsupervised domain adaptation, we are given a source domain $\mathcal{D}_s=\{(\mathbf{x}^s_i,y^s_i)\}^{n_s}_{i=1}$ of $n_s$ labeled examples where $y^s_i \in \{ 1, 2, \dots, C \}$ and a target domain $\mathcal{D}_t=\{ \mathbf{x}^t_j\}^{n_t}_{j=1}$ of $n_t$ unlabeled examples. The source domain and the target domain are sampled from different probability distributions $P$ and $Q$ respectively, and $P \neq Q$. The goal is to design a deep neural network $y=f(\mathbf{x})$ that formally reduces the shifts of the distributions across domains and enables learning multiple transferable representations, such that the target risk $R_t(f)=\mathbb{E}_{(\mathbf{x}, y)~Q}[f(\mathbf{x}) \neq y]$ can be minimized by minimizing the source risk and domain discrepancy.

In recent years, the deep transfer networks have achieved remarkable results~\cite{long2015learning,tzeng2017adversarial}. We call these methods as single-representation adaptation methods since they only align the distributions from representations extracted by a single structure. However, the single-representation adaptation methods focus on the partial information of the samples mentioned above, thus they might not work well for diverse scenarios. Comparing to single-representation methods which only contain partial information, multi-representation models might cover more information. In other words, we aim to learn multiple domain-invariant representations. Along this line, we propose a hybrid structure named Inception Adaptation Module (IAM) which contains multiple substructures to extract multiple representations from images.

To achieve MRA, it is necessary to minimize the discrepancy between the distributions of the multiple representations extracted from the source and target domains. To this end, Maximum mean discrepancy (MMD)~\cite{tzeng2014deep,long2015learning} is extended to conditional maximum mean discrepancy (CMMD) which could compute the discrepancy of conditional distributions for multiple representations. Based on IAM and CMMD, we propose multi-representation adaptation network (MRAN). Note that, different from previous methods minimizing the discrepancy between the distributions of single representation, MRAN can align the distributions of multiple representations.

%%%With the desire to minimize the discrepancy, Maximum mean discrepancy (MMD) has widely used as adaptation loss~\cite{tzeng2014deep, long2015learning}. However, comparing to previous one-representations adaptation methods which only minimize single adaptation loss, minimizing the discrepancy between the distributions of multiple representations extracted by IAM is more difficult. In order to compute the more complete multiple adaptation loss, we extend MMD to conditional maximum mean discrepancy (CMMD) which could could compute the discrepancy of conditional distributions~\cite{}, and use the CMMD as the adaptation loss.

\subsection{Inception Adaptation Module}
Similar structures are adopted for recent convolutional neural networks, e.g., ResNet~\cite{he2016deep}, DenseNet~\cite{huang2017densely}, and generally the structure $y = f(\mathbf{x})$ is divided into three parts $g(\cdot)$, $h(\cdot)$, $s(\cdot)$. The first part is the convolutional neural network $g(\cdot)$, which is used to convert high-pixel images to low-pixel ones; the second part $h(\cdot)$ is the global average pooling to extract representations from low-pixel images; the third part is the classifier $s(\cdot)$ to predict labels. Hence, $y = f(\mathbf{x})$ is reformulated as $y = (s \circ h \circ g)(\mathbf{x})$ ( $(h \circ g)(\mathbf{x}) = h( g (\mathbf{x}) )$ ).

Some recent deep transfer methods~\cite{long2016deep,pei2018multi} use the activations of the global average pooling layer as image representations and then align the distributions of the single representation. However, this single-representation adaptation manner might miss some important information for further performance improvement. Thus it is necessary to learn multiple domain-invariant representations by minimizing the discrepancy between the distributions of multiple representations.

\begin{figure}[t!]
\centering
\begin{minipage}[b]{1\linewidth}
\centering
\includegraphics[width=4.5in]{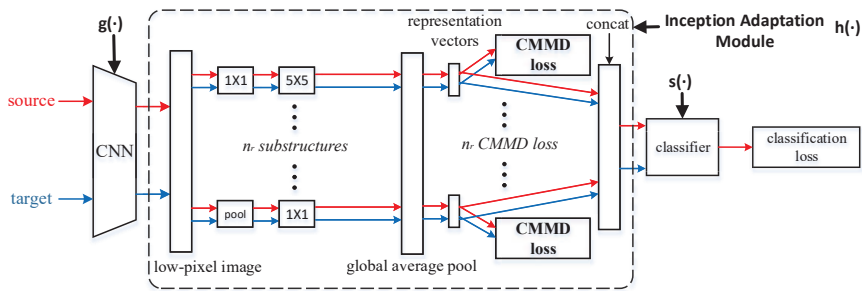}
\end{minipage}
\caption{Multi-Representation Adaptation Network (MRAN) aligns the conditional distributions of multiple representations. Inception Adaptation Module (IAM) could extract multiple representations from low-pixel images. By minimizing CMMD loss, the conditional distributions between the source and target domains are drawn close.}\label{network}
\end{figure}

To learn multiple different domain-invariant representations, the easiest way is to train multiple different convolutional neural networks. However, it would be very time-consuming to train multiple convolutional neural networks. It is well known that different structure could extract different representations from images. Hence, we use the a hybrid structure IAM consisted of multiple substructures to extract multiple representations from low-pixel images. As an intuitive example shown in Figure~\ref{network}, IAM has multiple substructures $h_1(\cdot), \dots, h_{n_r}(\cdot)$ ($n_r$ is the number of substructures), which are different from each other. With the IAM replacing the global average pooling, multiple representations $(h_1 \circ g)(\mathbf{X}), \dots, (h_{n_r} \circ g)(\mathbf{X})$ can be obtained. 
Comparing to the single representation, the multiple representations could cover more information. Hence, aligning the distributions of the multiple representations with more information could achieve better performance. 
The adaptation task could be achieved by minimizing the discrepancy of distributions based on the multiple representations:
\begin{equation}
\min_{f} \sum_i^{n_r} \hat{d}((h_i \circ g)(\mathbf{X}_s), (h_i \circ g)(\mathbf{X}_t) ),
\label{adaploss}
\end{equation}
where $\mathbf{X}$ is the set of $\mathbf{x}$ and $\hat{d}(\cdot, \cdot)$ is an estimator of discrepancy between two distributions. To achieve the classification task, the concatenated vectors $[(h_1 \circ g)(\mathbf{X}); \dots; (h_{n_r} \circ g)(\mathbf{X})]$ are put into the classifier $s(\cdot)$ which contains a fully connected layer and a softmax layer. The fully connected layer is mainly used to recombine the multiple representations, and the softmax layer is used to output the predicted labels. Finally, the neural network $y = f(\mathbf{x})$ with IAM is reformulated as:
\begin{equation}
y = f(\mathbf{x}) = s( [(h_1 \circ g)(\mathbf{X}); \dots; (h_{n_r} \circ g)(\mathbf{X})] ).
\end{equation}
Different from previous single-representation adaptation networks, the deep transfer networks with IAM are capable to learn multiple domain-invariant representations. The IAM is a multi-representation extractor. Moreover, the multiple domain-invariant representations can cover more information.
It is worth noting that the IAM can be implemented by most feed-forward models. When you implement IAM in other networks, you just replace the last average pooling layer with IAM.

\subsection{Conditional Maximum Mean Discrepancy}
To measure the Equation~\ref{adaploss}, a major issue is to choose a proper distance measure. We first introduce the non-parametric distance estimation Maximum Mean Discrepancy (MMD)~\cite{gretton2012kernel} which has been widely used to measure the discrepancy of marginal distributions:
\begin{equation}
\begin{split}
    \hat{d}_\mathcal{H}(\mathbf{X}_s,\mathbf{X}_t)=\left\| \frac{1}{n_s} \sum_{\mathbf{x}_i\in \mathcal{D}_{\mathbf{X}^s}}\phi (\mathbf{x}_i)-\frac{1}{n_t} \sum_{\mathbf{x}_j\in \mathcal{D}_{\mathbf{X}^t}}\phi (\mathbf{x}_j)\right\|^2_\mathcal{H}.
    \label{unbiased-mmd}
\end{split}
\end{equation}
By minimizing Equation~\ref{unbiased-mmd}, the marginal distributions between the source and target domains are drawn close.

According to~\cite{elhamifar2013sparse}, the data samples from the same class should lay on the same subspace, even if they belong to different domains. Hence, we reduce the difference in the conditional distributions instead of marginal distributions. Indeed, minimizing the discrepancy between the conditional distributions $P_s(y_s|\mathbf{x}_s)$ and $Q_t(y_t|\mathbf{x}_t)$ is crucial for robust distribution adaptation~\cite{sun2011two}. Unfortunately, it is nontrivial to match the conditional distributions, even by exploring sufficient statistics of the distributions, since there are no labeled data in the target domain, i.e., $Q(y_t,\mathbf{x}_t)$ cannot be modeled directly.

Fortunately, the output of the deep neural network $\hat{y}_i^t=f(\mathbf{x}_i^t)$ could be used as the pseudo label for data in target domain. Since the posterior probabilities $P(y_s|\mathbf{x}_s)$ and $Q(y_t|\mathbf{x}_t)$ are hard to represent~\cite{long2013transfer}, we resort to explore the sufficient statistics of class-conditional distributions $P(\mathbf{x}_s|y_s = c)$ and $Q(\mathbf{x}_t|y_t = c)$ instead w.r.t. each class $c\in \{1,\dots,C\}$. Now with the true labels of source domain data and pseudo labels of target domain data, we can essentially match the class-conditional distributions $P(\mathbf{x}_s|y_s = c)$ and $Q(\mathbf{x}_t|y_t = c)$. Here we modify MMD to measure the distance between the class-conditional distributions  $P(\mathbf{x}_s|y_s = c)$ and $Q(\mathbf{x}_t|y_t = c)$, called CMMD:
\begin{equation}
\begin{split}
    \hat{d}_\mathcal{H}(\mathbf{X}_s,\mathbf{X}_t)=\frac{1}{C} \sum^C_{c=1} \left\| \frac{1}{n_s^{(c)}} \sum_{\mathbf{x}_i^{s(c)}\in \mathcal{D}_{\mathbf{X}^s}^{(c)}}\phi (\mathbf{x}_i^{s(c)})-\frac{1}{n_t^{(c)}} \sum_{\mathbf{x}_j^{t(c)}\in \mathcal{D}_{\mathbf{X}^t}^{(c)}}\phi (\mathbf{x}_j^{t(c)})\right\|^2_\mathcal{H}.
    \label{cmmd}
\end{split}
\end{equation}
By minimizing Equation~\ref{cmmd}, the conditional distributions between the source and target domains are drawn close. Though we adopt the pseudo labels of the target domain, we expect to iteratively improve the labeling quality during the optimization.
%
%If we use $\hat{y}^t$ as the pseudo target labels and run back-propagation iteratively, then we can iteratively improve the labeling quality until convergence.
%% This EM-like pseudo label refinement procedure is empirically effective.
\subsection{Multi-Representation Adaptation Network}
To enable effective unsupervised domain adaptation, we propose Multi-Representation Adaptation Network (MRAN) as shown in Figure~\ref{network}, which aligns the distributions of multiple representations extracted by IAM in an end-to-end deep learning model. Note that the features in the lower layers of the network are transferable and hence will not require a further distribution matching~\cite{yosinski2014transferable}. The loss of MRAN is formulated as:
\begin{equation}
\min_{f} \frac{1}{n_s} \sum^{n_s}_{i=1} J(f( \mathbf{x}^s_i ), \mathbf{y}^s_i) + \lambda \sum_i^{n_r} \hat{d}( (h_i \circ g)(\mathbf{X}_s), (h_i \circ g)(\mathbf{X}_t) ),
\end{equation}
where $J(\cdot,\cdot)$ is the cross-entropy loss function (classification loss), $\hat{d}(\cdot,\cdot)$ is domain adaptation loss calculated by Equation~\ref{cmmd}, and $\lambda > 0$ is the trade-off parameter. We implement MRAN based on ResNet and replace the global average pooling by IAM. Specifically, these layers in the network are tailored to task-specific structures, which are adapted by minimizing classification error and CMMD.

Note that, training deep CNN requires a large amount of labeled data, which is prohibitive for many domain adaptation applications, so we start with the CNN pre-training on ImageNet2012 data and fine-tune it as~\cite{long2016deep}. The training of MRAN mainly follows standard mini-batch stochastic gradient descent(SGD) algorithm. In each mini-batch, we sample the same number of source domain data and target domain data to eliminate the bias caused by domain size. 
\section{Experiments}\label{experiment}
We evaluate the Multi-Representation Adaptation Network (MRAN) against state-of-the-art domain adaptation methods on three datasets: \textbf{ImageCLEF-DA}, \textbf{Office-31} and \textbf{Office-Home}.

\subsection{Experimental setup}
\subsubsection{Datasets}
\textbf{ImageCLEF-DA}\footnote{http://imageclef.org/2014/adaptation.} is a benchmark dataset for ImageCLEF 2014 domain adaptation challenge, which is organized by selecting the 12 common categories shared by the following three public datasets, each is considered as a domain: Caltech-256 (C), ImageNet ILSVRC 2012 (I), and Pascal VOC 2012 (P). There are 50 images in each category and 600 images in each domain. We use all domain combinations and build 6 transfer tasks: I $\rightarrow$ P, P $\rightarrow$ I, I $\rightarrow$ C, C $\rightarrow$ I, C $\rightarrow$ P, P $\rightarrow$ C.

\textbf{Office-31}~\cite{saenko2010adapting} is a benchmark for domain adaptation, comprising 4,110 images in 31 classes collected from three distinct domains: Amazon(A), which contains images downloaded from amazon.com, Webcam(W) and DSLR(D), which contain images taken by web camera and digital SLR camera with different photographical settings. The images in each domain are unbalanced across the 31 classes. To enable unbiased evaluation, we evaluate all methods on all six transfer tasks A $\rightarrow$ W, D $\rightarrow$ W, W $\rightarrow$ D, A $\rightarrow$ D, D $\rightarrow$ A, W $\rightarrow$ A as in~\cite{long2016deep,tzeng2014deep,ganin2015unsupervised}.

\textbf{Office-Home}~\cite{venkateswara2017deep} is a new dataset which consists 15,588 images larger than Office-31 and ImageCLEF-DA. It consists of images from 4 different domains: Artistic images (A), Clip Art (C), Product images (P) and Real-World images (R). For each domain, the dataset contains images of 65 object categories collected in office and home settings.

\subsubsection{Baselines}
We compare MRAN with various kinds of competitors, including Transfer Component Analysis (TCA)~\cite{pan2011domain}, Geodesic Flow Kernel (GFK)~\cite{gong2012geodesic}, Deep Convolutional Neural Network ResNet~\cite{he2016deep}, Deep Domain Confusion (DDC)~\cite{tzeng2014deep}, Deep Adaptation Network (DAN)~\cite{long2015learning}, Deep CORAL (D-CORAL)~\cite{sun2016deep}, Reverse Gradient (RevGrad)~\cite{ganin2015unsupervised}, Joint Adaptation Networks (JAN)~\cite{long2016deep}, Multi-Adversarial Domain Adaptation (MADA)~\cite{pei2018multi} and Collaborative and Adversarial Network (CAN)~\cite{zhang2018collaborative}.

To further validate the effectiveness of conditional distribution adaptation and IAM, we also evaluate several variants of MRAN: (1) MRAN (CMMD), which adds the CMMD module to ResNet; (2) MRAN (IAM), which uses IAM without adaptation loss; (3) MRAN (CMMD+IAM), which uses IAM with CMMD as the adaptation loss. Note that MRAN (CMMD) improves DAN~\cite{long2015learning} by replacing the multiple MMD penalties in DAN by the CMMD penalty. Besides, MRAN (IAM) imporves ResNet~\cite{he2016deep} by replacing the global average pooling layers by IAM. Inspired by GoogLeNet~\cite{szegedy2015going}, we use four substructures ($n_r=4$) in this work. However, you can set any number of substructures for other applications. (substructure1: conv1$\times$1, conv5$\times$5; substructure2: conv1$\times$1, conv3$\times$3, conv3$\times$3; substructure3: conv1$\times$1; substructure4: pool, conv1$\times$1).
% In the results, the MRAN means MRAN (cmmd+IAM).

\subsubsection{Implementation Details}
We employ ResNet~(50 layers) to learn transferable deep representations and use the activations of the last feature layer $pool5$ as image representation for baselines~\cite{long2016deep}. Following standard evaluation protocols for unsupervised domain adaptation~\cite{long2015learning,ganin2015unsupervised}, we use all labeled source examples as the source domain and all unlabeled target examples as the target domain. The average classification accuracy and standard error over three random trials are reported for comparison. For all baseline methods, we either follow their original model selection procedures or conduct transfer cross-validation~\cite{zhong2010cross} if their model selection strategies are not specified. For MMD-based methods (TCA, DDC, DAN, RTN, JAN, MRAN), we adopt Gaussian kernel with bandwidth set to median pairwise squared distances on the training data~\cite{gretton2012kernel}.

All deep methods are implemented base on the pytorch framework, and fine-tune from pytorch-provided models of ResNet~\cite{he2016deep}. We fine-tune all convolutional and pooling layers and train the classifier layer via back propagation. Since the classifier is trained from scratch, we set its learning rate to be 10 times that of the other layers. We use mini-batch stochastic gradient descent (SGD) with momentum of 0.9 and the learning rate annealing strategy in RevGrad~\cite{ganin2015unsupervised}: the learning rate is not selected by a grid search due to high computational cost, it is adjusted during SGD using the following formula: ${\eta}_p = \frac{\eta_0}{(1+\alpha p)^\beta}$, where $p$ is the training progress linearly changing from $0$ to $1$, $\eta_0 = 0.01$, $\alpha = 10$ and $\beta = 0.75$, which is optimized to promote convergence and low error on the source domain. To suppress noisy activations at the early stages of training, instead of fixing the adaptation factor $\lambda$, we gradually change it from $0$ to $1$ by a progressive schedule: $\lambda_p = \frac{2}{exp(-\gamma p)} - 1$, and $\gamma = 10$ is fixed throughout the experiments~\cite{ganin2015unsupervised}. This progressive strategy significantly stabilizes parameter sensitivity and eases model selection for MRAN.

\subsection{Results}\label{results}
All the results of three datasets are shown in Tables~\ref{tab:Image-CLEF}, ~\ref{tab:office31} and~\ref{tab:officehome}, respectively. From these results, we have the following insightful observations:

$\bullet$ MRAN (CMMD+IAM) outperforms all comparison methods on most transfer tasks. Particularly, MRAN (CMMD+IAM) substantially improves the accuracy by large margins on ImageCLEF-DA dataset, which have the same number of images in different domains and different classes. The encouraging results indicate the importance of incorporating CMMD and IAM and validate that MRAN (CMMD+IAM) is able to learn better transferable representations.

$\bullet$ Comparing DAN with MRAN (CMMD) with the same Gaussian kernel, the only difference is that MRAN (CMMD) aligns the conditional distributions, while DAN aligns the marginal distributions. MRAN (CMMD) is better than DAN, and the reason may be that the data samples from the same category should lay in the same subspace, even if they belong to different domains~\cite{elhamifar2013sparse}.

$\bullet$ MRAN (CMMD + IAM) substantially outperforms MRAN (CMMD), which shows the importance of aligning distributions of multiple representations rather than single representation.

$\bullet$ MRAN (CMMD+IAM) performs better than MRAN (IAM) while other deep transfer learning methods perform better than ResNet, which indicates the importance of transfer learning.

Note that, different from all the previous deep transfer learning methods that only align the marginal distributions of representations extracted by a single structure, while our model aligns the conditional distributions of multiple representations extracted by a hybrid structure (IAM), which implies that MRAN (CMMD+IAM) has a more powerful transferrable ability.

\begin{table}[!th]
\small
\centering
\caption{Accuracy(\%) on ImageCLEF-DA for unsuperevised domain adaptation (ResNet)} \label{tab:Image-CLEF}
\begin{tabular}{@{}cccccccc@{}}
\toprule
Method & I $\rightarrow$ P & P $\rightarrow$ I & I $\rightarrow$ C & C $\rightarrow$ I & C $\rightarrow$ P & P $\rightarrow$ C & Avg \\
\midrule
ResNet~\cite{he2016deep} & 74.8$\pm$0.3 & 83.9$\pm$0.1 & 91.5$\pm$0.3 & 78.0$\pm$0.2 & 65.5$\pm$0.3 & 91.2$\pm$0.3 & 80.7\\
DDC~\cite{tzeng2014deep} & 74.6$\pm$0.3 & 85.7$\pm$0.8 & 91.1$\pm$0.3 & 82.3$\pm$0.7 & 68.3$\pm$0.4 & 88.8$\pm$0.2 & 81.8 \\
DAN~\cite{long2015learning} & 75.0$\pm$0.4 & 86.2$\pm$0.2 & 93.3$\pm$0.2 & 84.1$\pm$0.4 & 69.8$\pm$0.4 & 91.3$\pm$0.4 & 83.3\\
RevGrad~\cite{ganin2015unsupervised} & 75.0$\pm$0.6 & 86.0$\pm$0.3 & \textbf{96.2}$\pm$0.4 & 87.0$\pm$0.5 & 74.3$\pm$0.5 & 91.5$\pm$0.6 & 85.0\\
D-CORAL~\cite{sun2016deep} & 76.9$\pm$0.2 & 88.5$\pm$0.3 & 93.6$\pm$0.3 & 86.8$\pm$0.6 & 74.0$\pm$0.3 & 91.6$\pm$0.3 & 85.2\\
JAN~\cite{long2016deep} & 76.8$\pm$0.4 & 88.0$\pm$0.2 & 94.7$\pm$0.2 & 89.5$\pm$0.3 & 74.2$\pm$0.3 & 91.7$\pm$0.3 & 85.8\\
MADA~\cite{pei2018multi} & 75.0$\pm$0.3 & 87.9$\pm$0.2 & 96.0$\pm$0.3 & 88.8$\pm$0.3 & 75.2$\pm$0.2 & 92.2$\pm$0.3 & 85.8\\
CAN~\cite{zhang2018collaborative} & 78.2 & 87.5 & 94.2 & 89.5 & 75.8 & 89.2 & 85.8\\
\midrule
%DSANrpool & 78.7$\pm$0.3 & 90.8$\pm$0.3 & 94.5$\pm$0.1 & 89.1$\pm$0.2 & 75.5$\pm$0.4 & 93.3$\pm$0.3 & 87.0\\
MRAN (IAM) & 76.2$\pm$0.7 & 88.4$\pm$0.5 & 91.4$\pm$0.2 & 84.2$\pm$0.1 & 69.2$\pm$0.2 & 88.6$\pm$0.3 & 83.0\\
%MRAN (IAM) & 77.7$\pm$0.4 & 91.3$\pm$0.5 & 92.5$\pm$0.1 & 87.3$\pm$0.9 & 71.5$\pm$0.4 & 92.1$\pm$0.2 & 85.4\\
MRAN (CMMD) & 78.7$\pm$0.2 & 91.1$\pm$0.2 & 94.2$\pm$0.4 & 88.9$\pm$0.1 & 75.1$\pm$0.3 & \textbf{93.1}$\pm$0.1 & 86.9\\
MRAN (CMMD+IAM) & \textbf{78.8}$\pm$0.3 & \textbf{91.7}$\pm$0.4 & 95.0$\pm$0.5 & \textbf{93.5}$\pm$0.4 & \textbf{77.7}$\pm$0.5 & \textbf{93.1}$\pm$0.3 & \textbf{88.3}\\
%Inceptionsub & 78.8$\pm$0.1 & 93.1$\pm$0.9 & 95.2$\pm$0.1 & 93.2$\pm$0.4 & 77.3$\pm$0.3 & 93.4$\pm$0.2 & 88.5\\
\bottomrule
\end{tabular}
\end{table}

\begin{table}[!th]
\small
\centering
\caption{Accuracy(\%) on Office-31 for unsuperevised domain adaptation (ResNet)} \label{tab:office31}
\begin{tabular}{@{}cccccccc@{}}
\toprule
Method & A $\rightarrow$ W & D $\rightarrow$ W & W $\rightarrow$ D & A $\rightarrow$ D & D $\rightarrow$ A & W $\rightarrow$ A & Avg \\
\midrule
ResNet~\cite{he2016deep} & 68.4$\pm$0.5 & 96.7$\pm$0.5 & 99.3$\pm$0.1 & 68.9$\pm$0.2 & 62.5$\pm$0.3 & 60.7$\pm$0.3 & 76.1\\
TCA~\cite{pan2011domain} & 74.7$\pm$0.0 & 96.7$\pm$0.0 & 99.6$\pm$0.0 & 76.1$\pm$0.0 & 63.7$\pm$0.0 & 62.9$\pm$0.0  & 79.3\\
GFK~\cite{gong2012geodesic} & 74.8$\pm$0.0 & 95.0$\pm$0.0 & 98.2$\pm$0.0 & 76.5$\pm$0.0 & 65.4$\pm$0.0 & 63.0$\pm$0.0  & 78.8\\
DDC~\cite{tzeng2014deep} & 75.8$\pm$0.2 & 95.0$\pm$0.2 & 98.2$\pm$0.1 & 77.5$\pm$0.3 & 67.4$\pm$0.4 & 64.0$\pm$0.5 & 79.7\\
DAN~\cite{long2015learning} & 83.8$\pm$0.4 & 96.8$\pm$0.2 & 99.5$\pm$0.1 & 78.4$\pm$0.2 & 66.7$\pm$0.3 & 62.7$\pm$0.2 & 81.3\\
D-CORAL~\cite{sun2016deep} & 77.7$\pm$0.3 & 97.6$\pm$0.2 & 99.7$\pm$0.1 & 81.1$\pm$0.4 & 64.6$\pm$0.3 & 64.0$\pm$0.4 & 80.8\\
RevGrad~\cite{ganin2015unsupervised} & 82.0$\pm$0.4 & 96.9$\pm$0.2 & 99.1$\pm$0.1 & 79.7$\pm$0.4 & 68.2$\pm$0.4 & 67.4$\pm$0.5 & 82.2\\
JAN~\cite{long2016deep} & 85.4$\pm$0.3 & 97.4$\pm$0.2 & 99.8$\pm$0.2 & 84.7$\pm$0.3 & 68.6$\pm$0.3 & 70.0$\pm$0.4 & 84.3\\
MADA~\cite{pei2018multi} & 90.0$\pm$0.1 & 97.4$\pm$0.1 & 99.6$\pm$0.1 & \textbf{87.8}$\pm$0.2 & \textbf{70.3}$\pm$0.3 & 66.4$\pm$0.3 & 85.2\\
CAN~\cite{zhang2018collaborative} & 81.5 & \textbf{98.2} & 99.7 & 85.5 & 65.9 & 63.4 & 82.4\\
\midrule
MRAN (IAM) & 77.4$\pm$0.5 & 96.1$\pm$0.5 & 99.5$\pm$0.1 & 81.9$\pm$0.8 & 64.2$\pm$0.4 & 64.8$\pm$0.9 & 80.7\\
MRAN (CMMD) & 87.0$\pm$0.4 & 97.7$\pm$0.2 & \textbf{100.0}$\pm$0.0 & 85.8$\pm$0.5 & 67.3$\pm$0.1 & 66.2$\pm$0.2 & 84.0\\
MRAN (CMMD+IAM) & \textbf{91.4}$\pm$0.1 & 96.9$\pm$0.3 & 99.8$\pm$0.2 & 86.4$\pm$0.6 & 68.3$\pm$0.5 & \textbf{70.9}$\pm$0.6 & \textbf{85.6}\\
%Inceptionsub & 92.3$\pm$0.8 & 97.2$\pm$0.2 & 99.9$\pm$0.1 & 88.7$\pm$0.9 & 68.2$\pm$0.8 & 71.5$\pm$0.6 & 86.3\\
\bottomrule
\end{tabular}
\end{table}

\begin{table}[!th]
\small
\centering
\caption{Accuracy(\%) on Office-Home for unsuperevised domain adaptation (ResNet)} \label{tab:officehome}
\begin{tabularx}{\textwidth}{cXXXXXXXXXXXXX}
\toprule
Method & A$\rightarrow$C & A$\rightarrow$P & A$\rightarrow$R & C$\rightarrow$A & C$\rightarrow$P & C$\rightarrow$R & P$\rightarrow$A & P$\rightarrow$C & P$\rightarrow$R & R$\rightarrow$A & R$\rightarrow$C & R$\rightarrow$P & Avg \\
\midrule
ResNet~\cite{he2016deep} & 48.5 & 68.3 & 75.4 & 53.8 & 64.4 & 66.1 & 52.7 & 42.8 & 74.1 & 65.3 & 49.6 & 79.7 & 61.1\\
DDC~\cite{tzeng2014deep} & 50.5 & 66.5 & 75.0 & 53.6 & 62.6 & 65.1 & 53.2 & 44.8 & 73.7 & 64.1 & 50.8 & 78.2 & 61.5\\
D-CORAL~\cite{sun2016deep} & 51.5 & \textbf{68.9} & \textbf{76.3} & 55.8 & 65.1 & 67.2 & 54.7 & 45.3 & 75.2 & 67.0 & 53.6 & 80.3 & 63.4\\
DAN~\cite{long2015learning} & 53.3 & 68.8 & 75.9 & 56.9 & 64.8 & 66.5 & 56.0 & 49.7 & 75.0 & 68.2 & 56.5 & 80.3 & 64.3\\
JAN~\cite{long2016deep} & 52.6 & \textbf{68.9} & \textbf{76.3} & \textbf{57.7} & 66.0 & 67.6 & 56.3 & 48.5 & 76.0 & 68.1 & 55.7 & 81.2 & 64.6\\
\midrule
%DSANpool & 53.0 & 68.6 & 76.2 & 59.4 & 66.2 & 67.6 & 58.2 & 52.4 & 76.1 & 68.5 & 56.7 & 81.1 & 65.3\\
MRAN (IAM) & 49.6 & 68.3 & 75.0 & 51.1 & 62.6 & 64.3 & 53.4 & 43.6 & 73.9 & 65.2 & 50.7 & 79.1 & 61.4\\
MRAN (CMMD) & 53.5 & 68.7 & 76.1 & 57.5 & 66.1 & 68.2 & 57.3 & 51.4 & 75.9 & 68.3 & 57.4 & 81.1 & 65.1\\
MRAN (CMMD+IAM) & \textbf{53.8} & 68.6 & 75.0 & 57.3 & \textbf{68.5} & \textbf{68.3} & \textbf{58.5} & \textbf{54.6} & \textbf{77.5} & \textbf{70.4} & \textbf{60.0} & \textbf{82.2} & \textbf{66.2}\\
%DSANfc & atoc & atop & ator & ctoa & ctop & ctor & ptoa & ptoc & ptor & rtoa & rtoc & rtop & \\
\bottomrule
\end{tabularx}
\end{table}

\subsection{Analysis}
To study the learnt representations of our model, we use MRAN (r1) as the representations extracted by the substructure1 (conv1$\times$1, conv5$\times$5), MRAN (r2) as the representations extracted by the substructure2 (conv1$\times$1, conv3$\times$3, conv3$\times$3), MRAN (r3) as the representations extracted by the substructure3 (conv1$\times$1) and MRAN (r4) as the representations extracted by the substructure4 (pool, conv1$\times$1). In addition, MRAN means the combined representations after fully connected layer.

\begin{figure}[t!]
\centering
\subfigure[MRAN]{
\begin{minipage}[b]{0.3\linewidth}
\centering
\includegraphics[width=1.5in,height=1.5in]{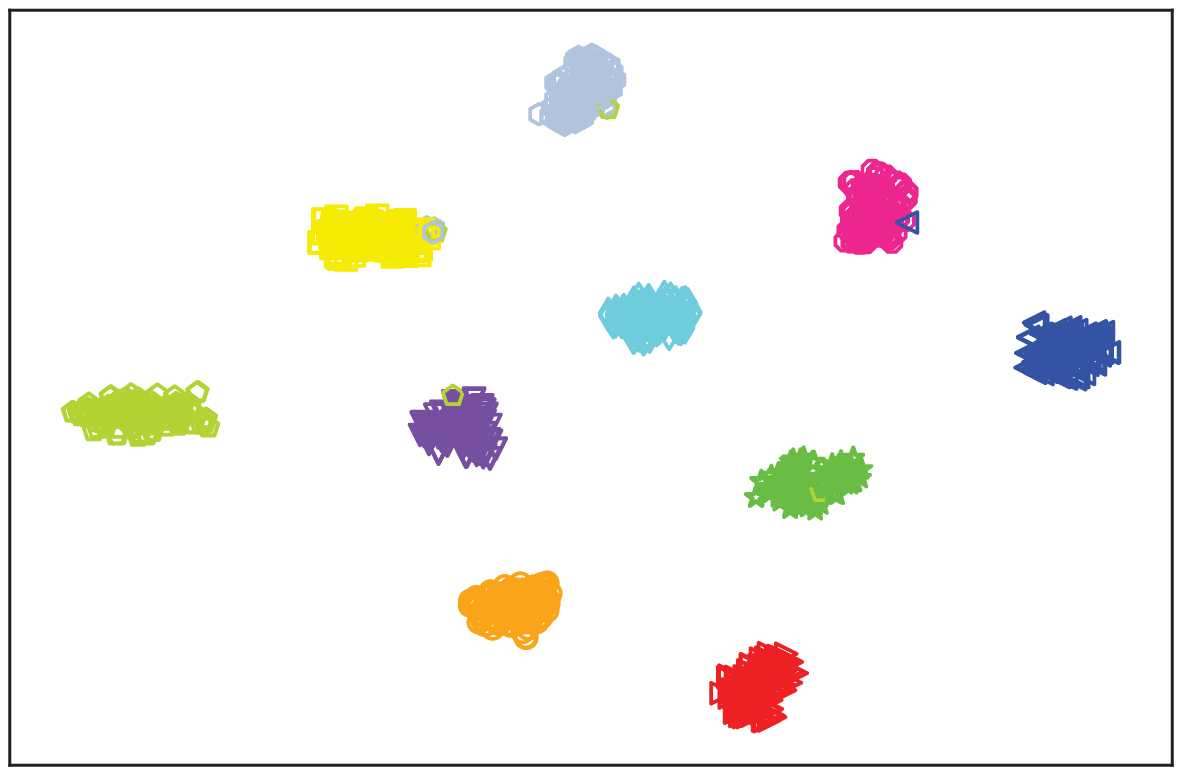}
\label{fig:4a}
\end{minipage}
}
\subfigure[MRAN(r1)]{
\begin{minipage}[b]{0.3\linewidth}
\centering
\includegraphics[width=1.5in,height=1.5in]{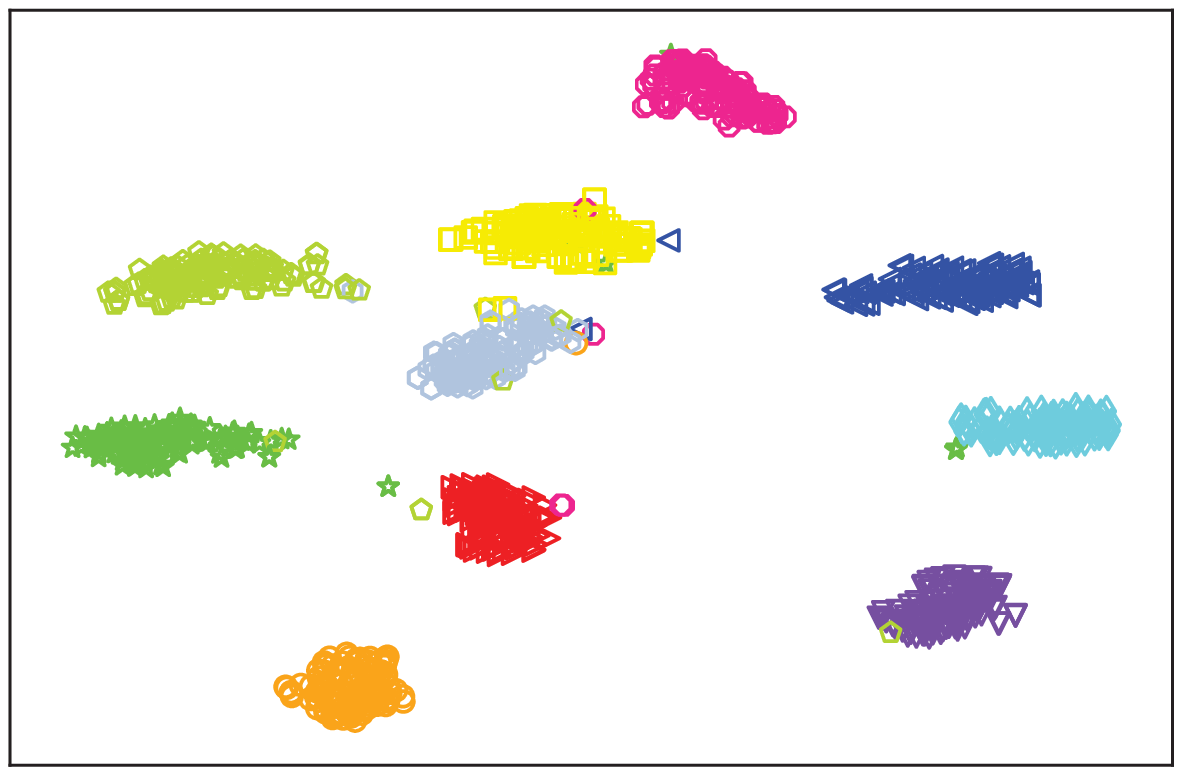}
\label{fig:4b}
\end{minipage}
}
\subfigure[MRAN(r2)]{
\begin{minipage}[b]{0.3\linewidth}
\centering
\includegraphics[width=1.5in,height=1.5in]{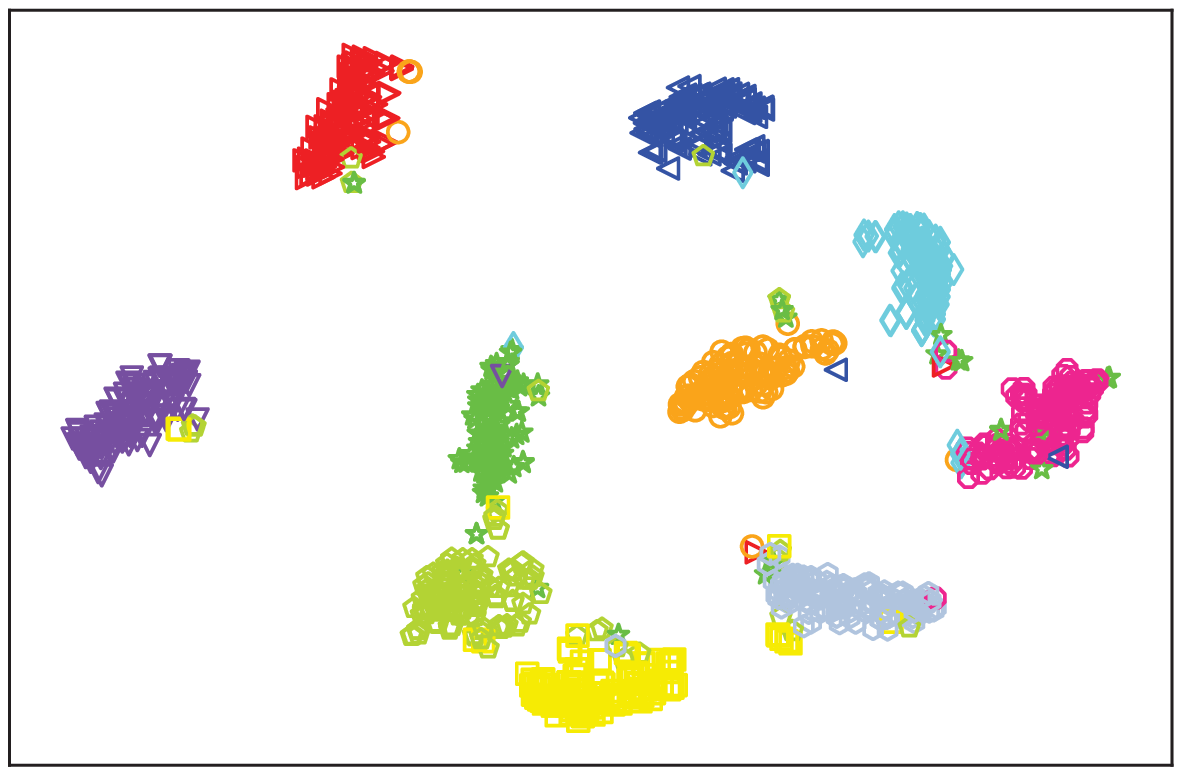}
\label{fig:4c}
\end{minipage}
}

\subfigure[MRAN(r3)]{
\begin{minipage}[b]{0.3\linewidth}
\centering
\includegraphics[width=1.5in,height=1.5in]{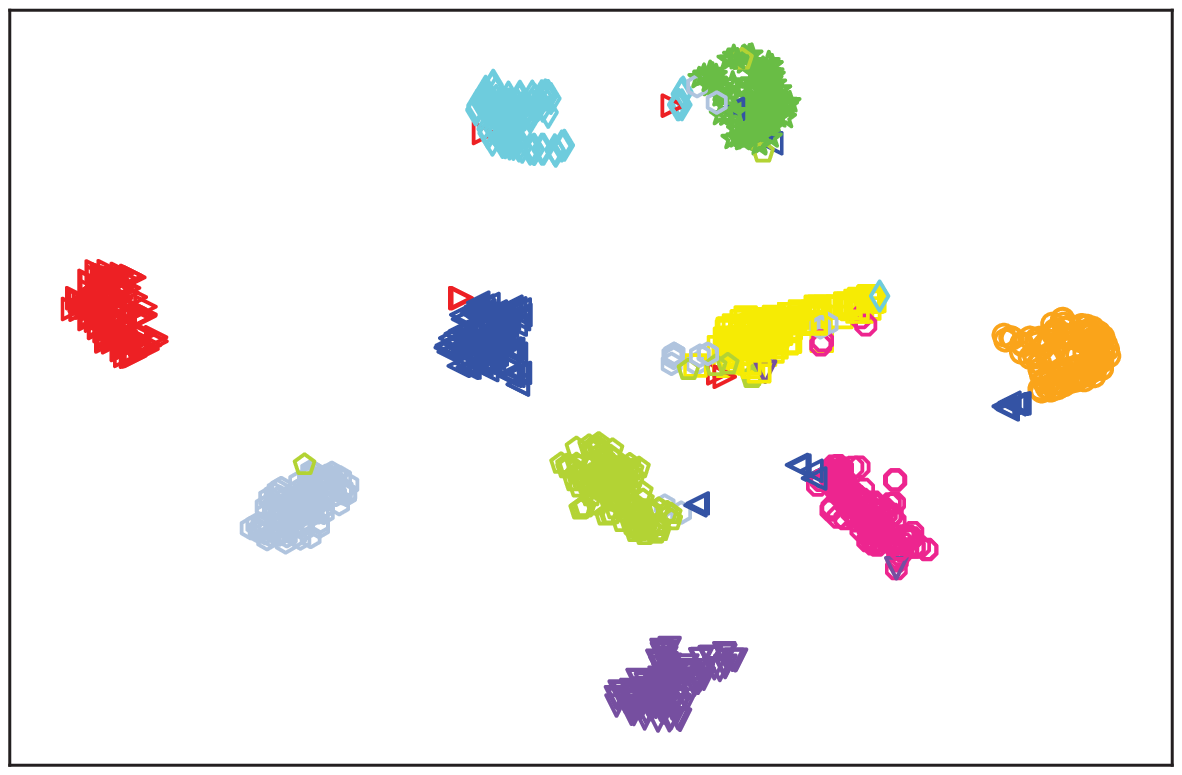}
\label{fig:4d}
\end{minipage}
}
\subfigure[MRAN(r4)]{
\begin{minipage}[b]{0.3\linewidth}
\centering
\includegraphics[width=1.5in,height=1.5in]{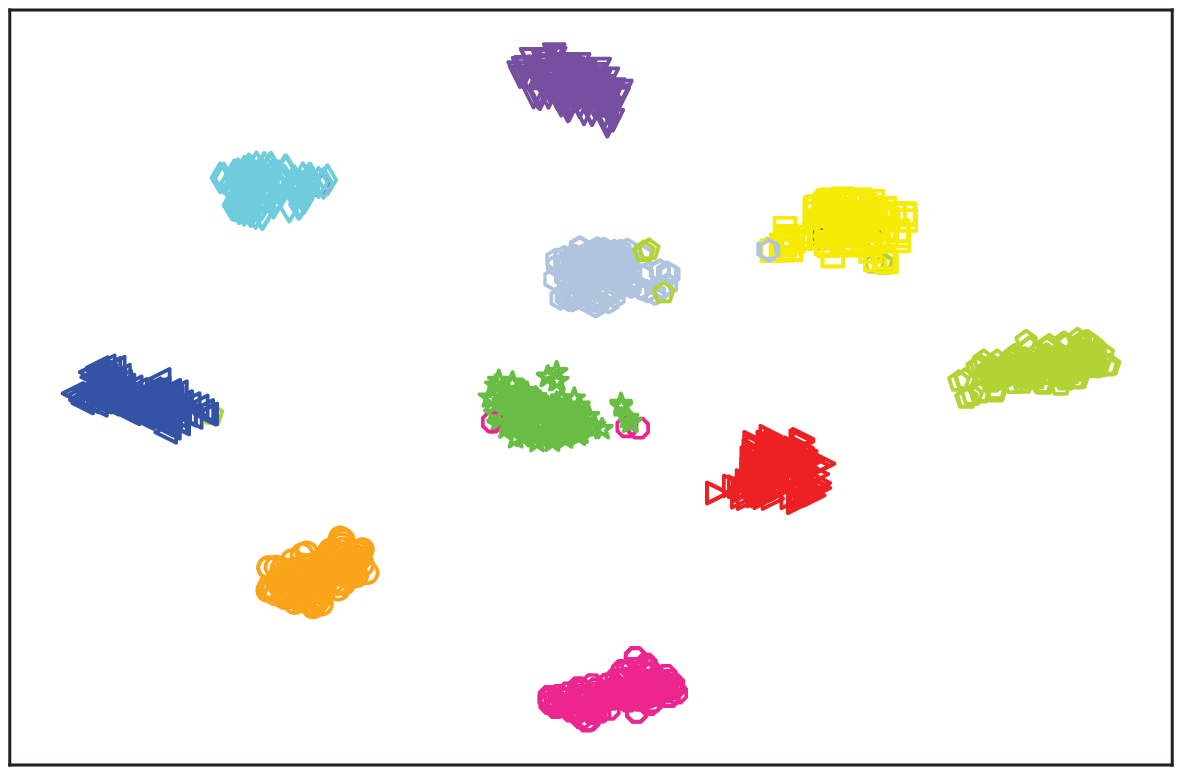}
\label{fig:4e}
\end{minipage}
}
\subfigure[DAN]{
\begin{minipage}[b]{0.3\linewidth}
\centering
\includegraphics[width=1.5in,height=1.5in]{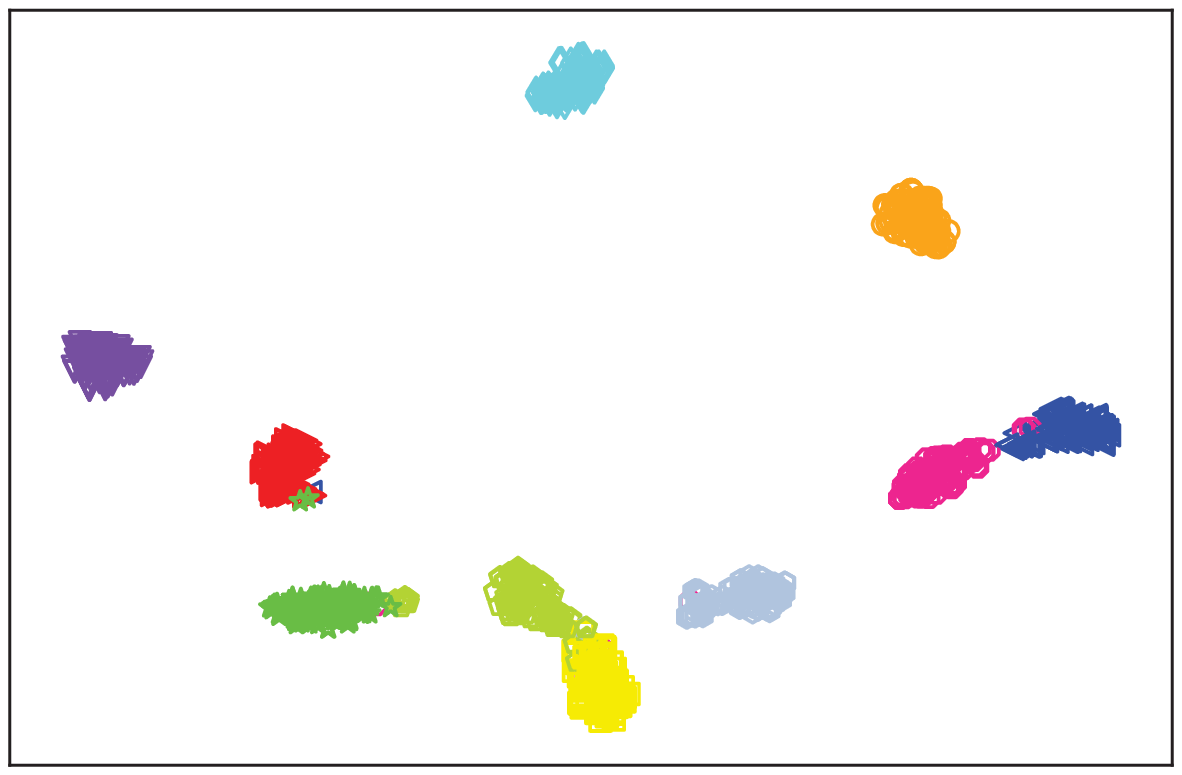}
\label{fig:5a}
\end{minipage}
}
\caption{Feature visualization: t-SNE of representations on source and target (a) MRAN; (b) MRAN (r1); (c) MRAN (r2); (d) MRAN (r3); (e) MRAN (r4); (f) DAN.}\label{fig:4}
\end{figure}

\begin{figure}[t!]
\centering
\subfigure[$\mathcal{A}$-distance]{
\begin{minipage}[b]{0.55\linewidth}
\centering
\includegraphics[width=2.5in, height=1.5in]{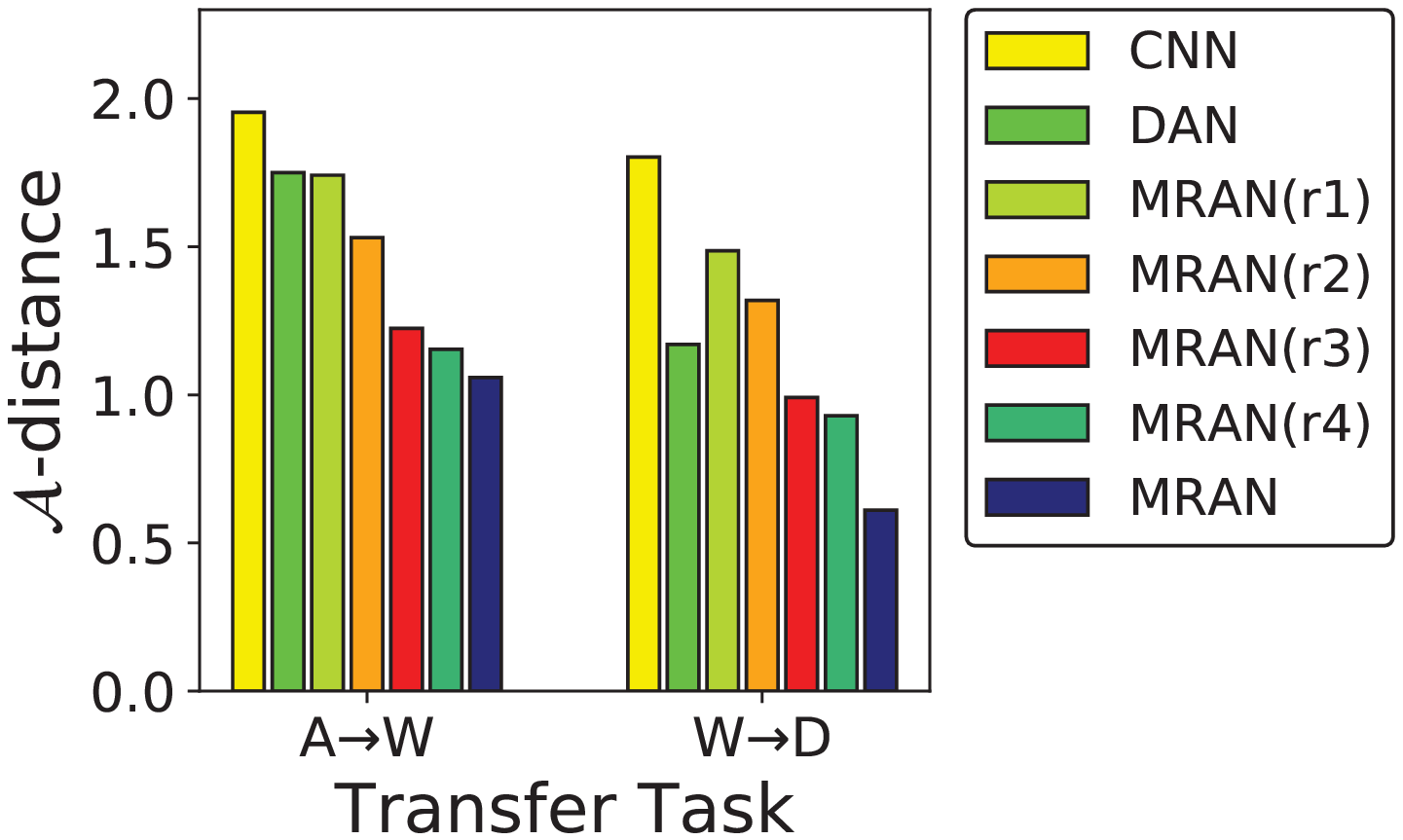}
\label{fig:5c}
\end{minipage}
}
\subfigure[Accuracy w.r.t $\lambda$]{
\begin{minipage}[b]{0.4\linewidth}
\centering
\includegraphics[width=1.8in,height=1.5in]{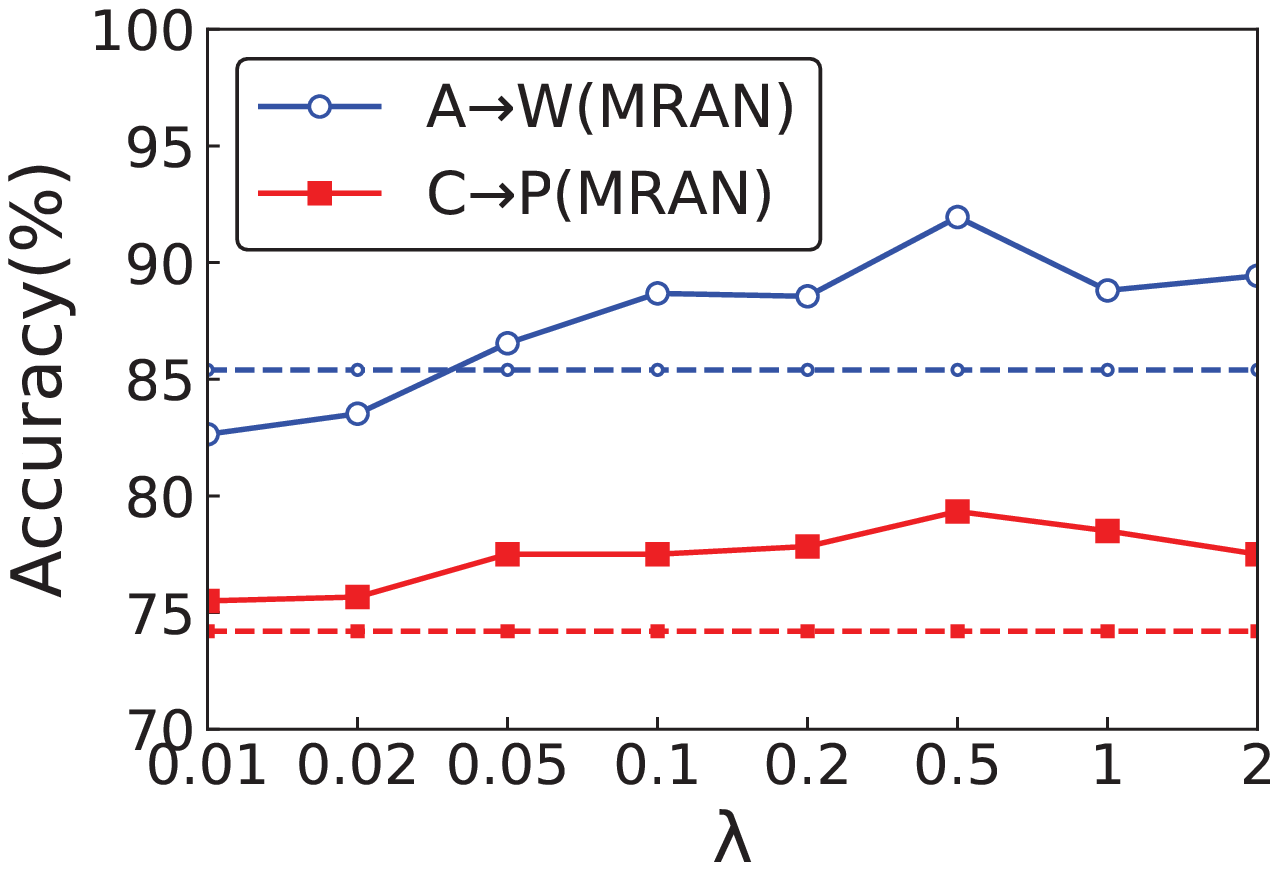}
\label{fig:5b}
\end{minipage}
}
\caption{(a) $\mathcal{A}$-distance; (b) parameter sensitivity of $\lambda$. }
\label{fig:5}
\end{figure}

\textbf{Feature Visualization: } We visualize the network representations of task A $\rightarrow$ W learned by MRAN and DAN using t-SNE embeddings~\cite{donahue2014decaf} in Figure~\ref{fig:4} and Figure~\ref{fig:5a}.

Comparing the representations given by MRAN (r1), MRAN (r2), MRAN (r3), MRAN (r4) in Figures~\ref{fig:4b}-~\ref{fig:4e}, the multiple representations extracted by different substructures have different distributions and a different number of wrong clustering. All of these demonstrate that different neural structures have the ability to extract different representations and the multiple representations have different information. Comparing with the representations given by MRAN and DAN in Figure~\ref{fig:4a} and Figure~\ref{fig:5a}, The combined representations given by MRAN in Figure~\ref{fig:4a} shows that the target categories are discriminated much more clearly.

\textbf{Distribution Discrepancy:} The domain adaptation theory~\cite{ben2010theory,mansour2009domain} suggests the distribution discrepancy measure $\mathcal{A}$-distance, together with the source risk, will bound the target risk. Specifically, the proxy $\mathcal{A}$-distance is defined as $d_\mathcal{A} = 2(1 - 2\epsilon)$, where $\epsilon$ is the generalization error of a classifier (e.g. kernel SVM) trained on the binary problem of discriminating the source and target domain data. Figure~\ref{fig:5c} shows the results of $d_\mathcal{A}$ on tasks A $\rightarrow$ W, W $\rightarrow$ D with learnt representations of CNN, DAN and MRAN. The $d_\mathcal{A}$ of the combined representations in Figure~\ref{fig:5c} are smaller than the $d_\mathcal{A}$ of CNN, DAN, MRAN (r1), MRAN (r2), MRAN (r3) and MRAN (r4). All of these results demonstrate that the combined representations from multiple representations are better transferable than single representation. In addition, these also prove that MRA which align distributions of multiple representations extracted by a hybrid structure could achieve better performance than previous single-representation adaptation methods.

%In addition, the $d_\mathcal{A}$ of multiple representations extracted by different substructures in Figure~\ref{fig:5c} is different.
%%\textbf{Ability of Different Substructures: }
%%\textbf{Effectiveness of the Multi-Representation Adaptation: } And

\textbf{Parameter Sensitivity:} We check the sensitivity of multiple adaptation loss parameter $\lambda$, which controls the relative weight for multiple adaptation loss. Figure~\ref{fig:5b} shows the performance of MRAN based on ResNet on tasks A $\rightarrow$ W and C $\rightarrow$ P by varying $\lambda \in \{0.01, 0.02, 0.05, 0.1, 0.2, 0.5, 1, 2 \}$. The accuracy of MRAN first increases and then decreases as $\lambda$ varies and displays as a bell-shaped curve. From the results, we can find a proper trade-off value about 0.5 to achieve good transfer performance. 

\textbf{Time Complexity:} CMMD loss involving pseudo label and the IAM indeed need some extra computations. We conduct additional experiments to record the time consuming of ResNet, DAN, MRAN(IAM) and MRAN(CMMD) for each iteration. All the experiments are conducted on a GeForce GTX 1080 Ti GPU, and the average time of each iteration over 100 iterations is recorded. The results are listed as follows: ResNet (0.147s), DAN (0.277s), MRAN(IAM) (0.173s) and MRAN(CMMD) (0.291s). Comparing MRAN(IAM) with ResNet, we could find that the IAM takes 0.025s more for each iteration which is very small. Comparing DAN with MRAN(CMMD), the only difference is that DAN uses MMD to align distributions and MRAN(CMMD) uses CMMD to align distributions. MRAN(CMMD) spends 0.014s more for each iteration than DAN, which is a common property of conditional alignment methods~\cite{pei2018multi}. Overall, though the use of IAM and CMMD would slightly increase the time complexity, it is reasonably and can greatly improve the performance.

\textbf{Insightful Findings:}
We get some findings from the experiments. (\textbf{1}) The IAM could extract multiple representations from images. The different representations could represent diverse abstract (underlying) views and we could see the t-SNE embeddings of different representations are different in Figure 3. (\textbf{2}) The MRA method (our MRAN) could achieve better performance than the single-representation adaptation methods as we could see in the Section 4.2, which also demonstrate the effectiveness of MRA. (3) Aligning conditional distributions are more effective than aligning marginal distributions.
\section{Conclusion}
\label{sec:conclusion}
%To overcome the problem of aligning distributions of only one single representation from the source and target domains, we tried to propose Multi-Representation Adaptation (MRA) to align the distributions of multiple representations. Along this line, we proposed Multi-Representation Adaptation (MRA) to learn multiple domain-invariant representations which might contain more information, and aligned them in a framework of Multi-Representation Adaptation Network (MRAN). In particularly, we proposed a hybrid neural structure named Inception Adaptation Module(IAM) to extract multiple representations from images. Note that, our framework can be adapted to different networks. Moreover, we extended the marginal distribution discrepancy measure MMD to conditional MMD (CMMD), which is effectively incorporated in our model. Finally, Extensive experiments are conducted on three datasets to demonstrate the effectiveness of the proposed model.

To overcome the problem of aligning distributions of only one single representation from the source and target domains, we tried to propose Multi-Representation Adaptation (MRA) to align the distributions of multiple representations. Along this line, we proposed a framework of Multi-Representation Adaptation Networks (MRAN) to learn multiple domain-invariant representations which might contain more information. In particularly, we proposed a hybrid neural structure named Inception Adaptation Module (IAM) to extract multiple representations from images. Note that, our framework can be adapted to different networks. Moreover, we extended the marginal distribution discrepancy measure MMD to conditional MMD, which is effectively incorporated in our model. Finally, Extensive experiments are conducted on three datasets to demonstrate the effectiveness of the proposed model.
\section{Acknowledgments}
\label{sec:ack}
The research work supported by the National Key Research and Development Program of China under Grant No. 2018YFB1004300, the National Natural Science Foundation of China under Grant No. U1836206, U1811461, 61773361, the Project of Youth Innovation Promotion Association CAS under Grant No. 2017146.

\bibliographystyle{splncs04}
\bibliography{main}
\end{document}